\documentclass[letterpaper, 10 pt, conference]{ieeeconf}

\IEEEoverridecommandlockouts
\usepackage{cite}
\usepackage{amsmath,amssymb,amsfonts}
\usepackage{algorithmic}
\usepackage{graphicx}
\usepackage{textcomp}
\def\BibTeX{{\rm B\kern-.05em{\sc i\kern-.025em b}\kern-.08em
    T\kern-.1667em\lower.7ex\hbox{E}\kern-.125emX}}

\usepackage[skip=1ex, font=small]{caption}

\usepackage{algorithm}
\usepackage{array}
\usepackage[caption=false,font=scriptsize,labelfont=sf,textfont=sf]{subfig}
\usepackage{stfloats}
\usepackage{url}
\usepackage{verbatim}
\usepackage{color}
\usepackage[normalem]{ulem}
\usepackage{multirow}
\usepackage{multicol}
\usepackage[table,xcdraw]{xcolor}
\title{\LARGE \bf Learning Failure Prevention Skills for Safe Robot Manipulation

}
\author{Abdullah Cihan Ak$^{1}$, Eren Erdal Aksoy$^{2}$ and Sanem Sariel$^{1}$%
\thanks{
This work is supported by a grant from the Scientific and Technological Research Council of Turkey (TUBITAK), Grant No. 119E-436.}%
\thanks{$^{1}$Artificial Intelligence and Robotics Laboratory, Faculty of Computer
and Informatics Engineering, Istanbul Technical University, Maslak, Turkey
\{akab, sariel\}@itu.edu.tr}%
\thanks{$^{2}$School of Information Technology, Center for Applied Intelligent Systems Research, Halmstad University, Halmstad, Sweden}%
}

\begin{document}
\maketitle
\thispagestyle{empty}
\pagestyle{empty}

\begin{abstract}
Robots are more capable of achieving manipulation tasks for everyday activities than before. But the safety of manipulation skills that robots employ is still an open problem. Considering all possible failures during skill learning increases the complexity of the process and restrains learning an optimal policy. Beyond that, in unstructured environments, it is not easy to enumerate all possible failures beforehand. In the context of safe skill manipulation, we reformulate skills as base and failure prevention skills where base skills aim at completing tasks and failure prevention skills focus on reducing the risk of failures to occur. Then, we propose a modular and hierarchical method for safe robot manipulation by augmenting base skills by learning failure prevention skills with reinforcement learning, forming a skill library to address different safety risks. Furthermore, a skill selection policy that considers estimated risks is used for the robot to select the best control policy for safe manipulation. Our experiments show that the proposed method achieves the given goal while ensuring safety by preventing failures. We also show that with the proposed method, skill learning is feasible, novel failures are easily adaptable, and our safe manipulation tools can be transferred to the real environment.  
\end{abstract}


\section{Introduction}

Robots that are used in domestic environments require manipulation skills to perform various manipulation tasks ~\cite{ersen}. Considering a kitchen environment, a robot can be assigned to cook various recipes such as soup or a cake. Primitive or compound motor skills such as stirring and pouring are needed to make these recipes. Existing learning methods enable learning such skills effectively\cite{Mnih2015,
Kalashnikov2018}.It is also crucial to ensure that these skills are executed safely. However, even well-designed skills are prone to fail in the real world due to wrong assumptions, perceptual errors, or changing environmental conditions \cite{inceoglu2020fino}. These may threaten the integrity of the workspace.  For example, a robot stirring soup may spill the soup or slide the pan out of the stove which can be harmful to the people nearby, its workspace, and itself. Therefore, the capability of the robot should not be limited to performing manipulation skills only, they also need to monitor/detect potential failures and prevent them for safety. For that purpose, we propose learning failure prevention skills augmenting the safety of robotic manipulation.

\begin{figure}[!t]
\centering
\includegraphics[width=3.4in]{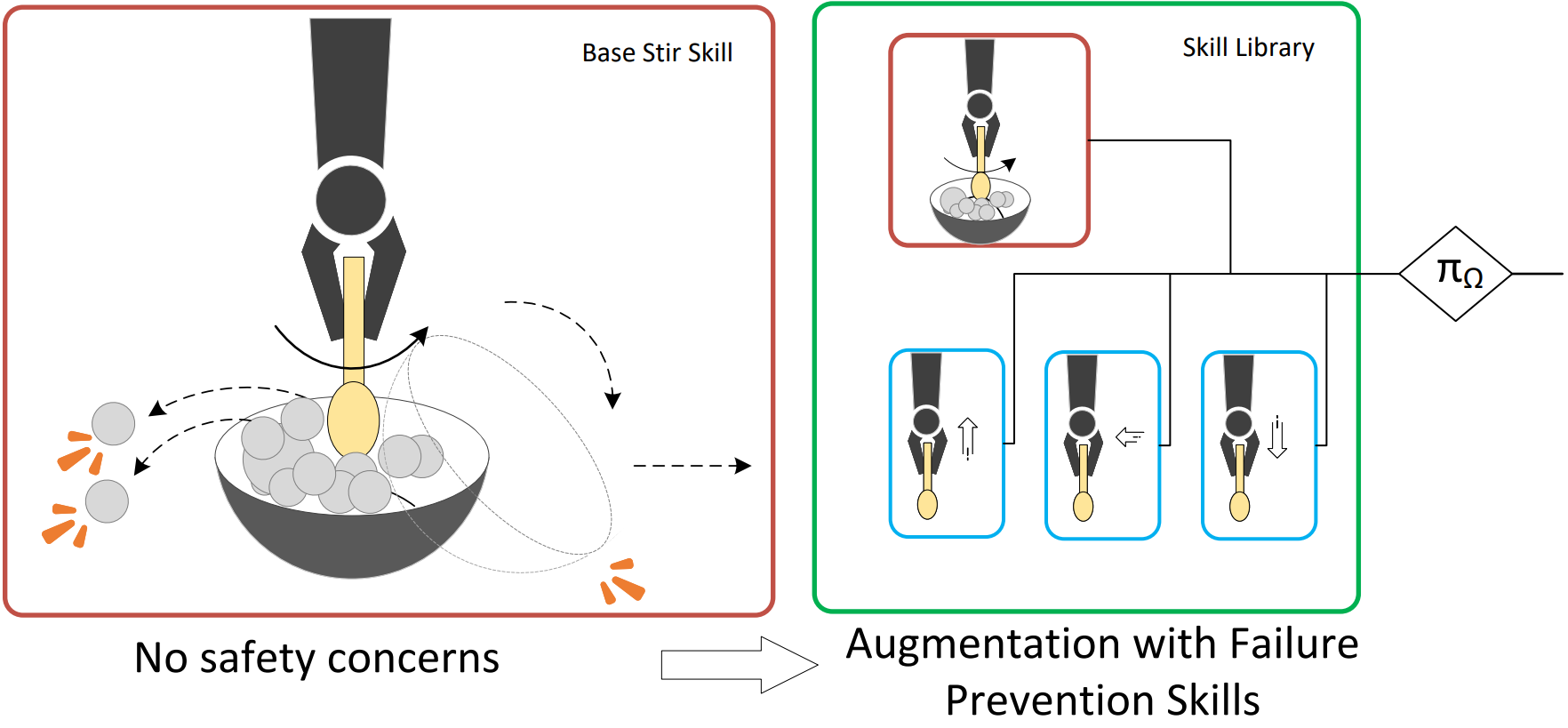}
\caption{A sample base skill, stir, may have been learned without taking into safety concerns. To make it safer, failure prevention skills are added in the library, and selected when needed.}
\label{fig:motivation}
\end{figure}

Manipulation skills are often considered closed-loop sensorimotor control systems for robots to complete various tasks. With recent developments in reinforcement learning (RL), even though robots can learn and use manipulation skills effectively for a limited number of objectives, they struggle when the number of objectives increases due to the curse of dimensionality. Therefore, RL-based skill learning rather focuses on achieving a given task without considering potential failures. Safe reinforcement learning approaches ~\cite{Bharadhwaj2020, Cheong2018, Pinto2017,Eysenbach2018} apply learning to satisfy the goal while keeping the execution safe by estimating the risk. These approaches have the intention to prevent failures by learning to avoid unsafe states, however, they do not respond to scenarios with several failures as in unstructured environments. A previous work ~\cite{ak2019stop} addresses this issue but with a limited number of failures. In summary, even though robots have better accuracy and persistence in manipulation tasks, humans surpass robots in detecting and preventing failures. Arguably, an optimal skill policy can complete the task without failures, however, it will be limited to the known failures which are experienced during training. For a novel failure, the previous skill would become obsolete and should be learned again. 

Representing a skill as smaller skills combined with a hierarchy ~\cite{Kroemer2021, Kroemer2015, Bacon_Harb_Precup_2017} benefits from reduced complexity of the goal for optimal skill learning. While such skills can be predefined, it is also possible to learn smaller skills incrementally ~\cite{Bagaria2020, Bagaria2021}. Even though hierarchical approaches focus on learning several skills none of the learned skills does not have an explicit focus on ensuring the safety of the execution.  For that purpose, the safety and the goal can be considered as two different objectives, and learning safe and robust skills can be expressed as a multi-objective reinforcement learning problem by decomposing the reward. Separating rewards and punishments can effectively perform both the task and failure prevention (\textit{contextual inhibition} and \textit{spontaneous recovery})~\cite{Lowe2013}. In the context of multi-objective reinforcement learning, positive rewards (reward) can be used to learn how to complete the task(main objective), whereas negative rewards (risk) can be introduced to avoid potential failures(safety objective). Many recent works ~\cite{VanSeijen2017, Lin2019, Lin2020, Elfwing2018, Wang2018} decompose the reward to learn to achieve the goal safely but they only respond to a single failure type. It can make a crucial difference to balance both rewards and risks whereas several objectives would result in suboptimal results for each objective. Especially, failures and their consequences are not limited and can vary in a dynamic scene, therefore, an increased number of objectives can make the multi-objective reinforcement learning problem unfeasible. 

To address these problems, we propose a modular method where base skills are augmented by failure prevention skills, enhancing their safety. We group skills in the skill library into two categories according to their purposes: base skills and failure prevention skills. Skills with the purpose of reaching a goal to complete a task are defined as base skills (i.e., stirring or pouring). Skills with the purpose of preventing failures define failure prevention skills (such as prevention of sliding, overturning, or spilling). The latter are more reusable skills for augmenting different base skills. The proposed method has a hierarchy between skills and skill selection. The lower level of the hierarchy is a skill library which is composed of base skills and failure prevention skills. For each potential failure, a risk estimation model is defined to estimate the risk of the failure happen in-near future.
Note that for the sake of simplicity, autonomous detection and identification of failures \cite{inceoglu2020fino} is out of the scope of this work. Failure prevention skills are learned and added to the skill library for preventing potential failures. The higher level in the hierarchy rather involves a skill selection policy, triggering the optimal skill from the skill library to safely accomplish the task. Figure \ref{fig:motivation} illustrates this concept.

In the case of detecting a novel failure, the proposed method can easily be revised by only learning a novel failure prevention skill that mitigates this particular new failure type. Therefore, it becomes easier to adapt to new environmental conditions where novel failures could be encountered.

The proposed method is evaluated in a simulated environment and the skill library formed in the simulated environment is transferred to a real environment for real world evaluation. To the best of our knowledge, this is the first study that addresses learning reusable failure prevention skills to enable failure precautions into a skill library.

\subsection{Contibutions}

Our main contributions are as follows: 
\begin{itemize}
    \item the formulation and implementation of a modular and hierarchical method for safe robot manipulation; 
    \item enabling learning reusable failure prevention skills as precautionary switching policies in a skill library for safe manipulation; 
    \item adapting to novel failures and augmenting incrementally;
    \item real-world applicability by effective safety precautions in the physical world.
\end{itemize}

\section{Learning Failure Prevention Skills for Safe Robot Manipulation}
\label{section:problem_description}

Cognitive robots are equipped with either hand-coded or learned motor skills to achieve a given task. Even though these skills work well for controlled environments, in unstructured environments, their outcomes are not always as expected, and even worse, undesired/unsafe situations may occur due to failures in changing situations. To ensure safety, it is crucial for the robot to anticipate potential failures before they occur, and to prevent them if possible. In this work, we address this problem and formulate it as augmenting base robot skills with appropriate precautions to make them safer without changing them.

\begin{figure}[!h]
\centering
\includegraphics[width=2.4in]{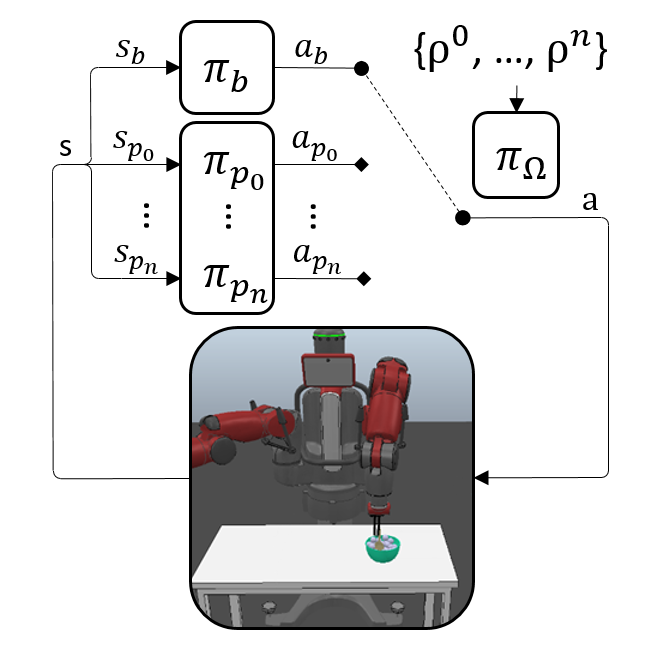}
\caption{Hierarchical Failure Prevention Model}
\label{fig:policy_selection}
\end{figure}

We define a base skill, $\pi_b$, as a motor skill to achieve either a primitive or a compound action (e.g., \textit {pick and place objects, pour liquids, stir a bowl of ingredients}, etc.). During performing this skill in a setting different than the trained one, failures are inevitable due to either wrong assumptions or unanticipated situations. The probability of a failure to occur is defined as a risk, ($\rho^k$, in the range from 0 and 1). The safety of the execution of $\pi_b$ can be monitored by continually checking the occurrence of a finite set of risks: $\{ \rho^0, ... \rho^n \}$ which can be designed previously or discovered online. The main problem that we address in this study asks; when the execution of a base skill $\pi_b$ increases the risk of a failure ($\rho^k$), to learn and activate necessary skills that enable transition to a safe state. We call this type of a complementary skill as failure prevention skill $\pi_{p_k}$ that is responsible for reducing a corresponding risk. Note that, these failure prevention skills are generic skills that can be used to augment different base skills. Therefore, the problem asks for learning these failure prevention skills, and taking over  control of  these skills ($\pi_{p_{0}}, ..., \pi_{p_{n}}$) when necessary during the execution of the base skill such that the whole execution is safe. Thus, a dynamic chain of skills (i.e., a linear sequence of the base skill and the failure prevention skills in different orders or by different selections) are executed based on the circumstances of the world. This problem also asks for effective and efficient selection of the prevention skills during execution.  

The skill library $L$ of the robot includes both the base skills and failure prevention skills (Equation \ref{eq:definition}). $L$ can be gradually built by adding novel skills online. Each learned skill extends the skill library to make it robust against each failure type that is observed.

\begin{equation}
\label{eq:definition}
L = \{\pi_{b_0},...,\pi_{b_m}\} \cup \{\pi_{p_0},...,\pi_{p_n}\}
\end{equation}

In our solution to this problem,  augmentation of a base skill  $\pi_b$ with safety precautions is done in three steps; observing failures and obtaining risk estimation models, learning failure prevention skills, and learning to select which skill to execute. In the first step, a failure is observed during the execution of the base skill $\pi_b$.  In the second step, a corresponding novel failure prevention skill $\pi_{k}$ is learned. In the third step, a skill selection policy $\pi_\Omega$ is used to select a skill from $L$ using the estimated risk models. The resulting model is depicted in Figure \ref{fig:policy_selection}. The following subsections describe this process in detail.

\subsection{Learning a Base Skill}
\label{sec:3behavior_skill_learning}

A base skill $\pi_b$ can be hand-coded by an expert or learned by optimizing models such as Markov Decision Process~(MDP) and Dynamic Movement Primitives~(DMP)~\cite{Ijspeert2013} using reinforcement learning~(RL) or learning from demonstration~(LfD). However, these models may not work as well as desired when the robot is exposed to reality as unexpected failures are likely in different settings other than the trained one. Therefore, a learned skill needs to be adapted to changes that occur in the environment. However, this adaptation may degrade the effectiveness of actual task performance (i.e., base skill) for the sake of adapting to changes. Therefore, we propose to augment the base skill to make it safer without changing it. This also ensures a more reusable solution to be used as a library of skills that can be used to augment other base skills as well.

\subsection{Risk Estimation Models}
\label{sec:3risk_estimation_model}

\begin{figure}[!b]
\centering
\includegraphics[width=1.5in]{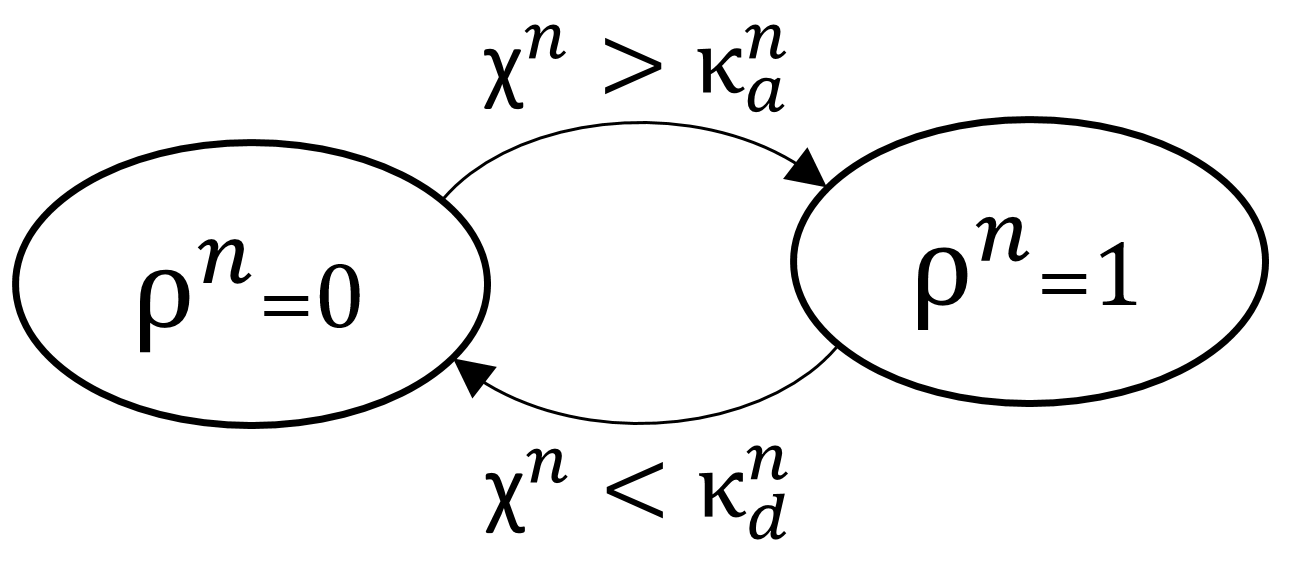}
\caption{Safe - Risky transition model}
\label{fig:risk_transition}
\end{figure}

A risk $\rho^n$ is a binary safety estimate  (\textit{safe} - \textit{risky}) against a failure, and presents the probability of a failure to occur in the near-future. In order to continue the execution safely, failure prevention decisions should be given based on risk estimations in real time.

Since the failure detection is not the main focus of this work, we use a rule-based risk estimation model which is expressed as a finite state machine $\rho^n = FSM\left(\chi^n, \kappa_a^n, \kappa_d^n\right)$ as illustrated in Figure \ref{fig:risk_transition}. A risk $\rho^n$ uses an observable parameter $\chi^n$ from the environment, and evaluates the risk using the activation threshold $\kappa_a^n$ and the deactivation threshold $\kappa_d^n$ where $\kappa_a^n\neq\kappa_d^n$. A safe state is transitioned to a risky state if $\chi^n$ is observed to be larger than $\kappa_a^n$ . A risky state is transitioned to a safe state if $\chi^n$ is observed to be smaller than $\kappa_d^n$. The interval between $\kappa_a^n$ and $\kappa_d^n$ prevents undesired fluctuations between states when the observed parameter is close to thresholds.

\subsection{Learning Failure Prevention Skills}
\label{sec:3prevention_skill_learning}
In our method, a failure prevention skill is modelled with MDP and learned to prevent a risky situation. The reward of a failure prevention skill is determined by the corresponding risk estimation as given in Equation \ref{eq:risk_reward}. Since risk estimations are binary, the reward is sparse. During the training, the failure prevention skill policy is optimized to maximize the reward as it minimizes the risk.
\begin{equation}
\label{eq:risk_reward}
    R^{risk} = 1 - \rho^{risk} 
\end{equation}
Risky states are expected to be observed occasionally. To learn to prevent risky states, the robot should observe these cases quite often. However, this is not the case if it is not  intended. In order to improve the sample efficiency of learning a failure prevention skill policy, the robot should be able to experience risky states more often. For that purpose, specific procedures are designed to create different failure situations. Therefore the robot uses these procedures to create a situation with a risk. Then the robot will be able to experience sequences of actions from risky situations to safety and learn an optimal policy that prevents failures.

\subsection{Skill Selection}
\label{sec:3skill_selection}
Once the skill library $L$ is formed, the robot can execute the task safely with a skill selection policy $\pi_\Omega$ that arbitrates over all skills in $L$. $\pi_\Omega$ achieves this selection in a  hierarchical fashion (See Figure \ref{fig:policy_selection}). $\pi_\Omega$ selects a skill from $L$ using risk estimations $\{\rho^0 ... \rho^n\}$, thus,  it is expected to select an appropriate failure prevention skill to reduce a risk to prevent any catastrophic situations before they occur, and select the base skill when appropriate to complete the task. 

We use a rule-based mapping  from risk estimations to skills in $L$ to form the skill selection policy $\pi_\Omega$. With  $\pi_\Omega$, the robot executes the base skill $\pi_b$ in a state without any risk. With the observation of a risk $\rho^k$, the corresponding failure prevention skill $\pi_{p_k}$ is selected to reduce the risk $\rho^k$. Upon detection of multiple risks, predefined priorities between failure prevention skills determine which skill to be executed first. Priorities are determined by the impact of the failure on the manipulation's safety reliability. Note that in risks are sorted from most important to least important, therefore, the failure prevention skill corresponding to the risk with the greatest index is selected first.


\section{Empirical Evaluation of Augmenting Stir Skill}
In this section, we present our case study on a continuous \textit{stirring} task. To accomplish this task, a Baxter humanoid robot uses a spoon to stir particles in a bowl. The goal is for the robot to move particles in the bowl for a given time. During the nominal execution of the \textit{stirring} skill, one natural failure is observed: \textit{spilling} the particles from the bowl. For different objects, environments, and/or control conditions two additional failures might occur in our setup, \textit{sliding} the bowl and \textit{overturning} the bowl. Failure prevention skills are learned and used to augment the \textit{stir} skill for the safety of the execution.

\begin{figure}[h]
\centering
\subfloat[]{\fbox{\includegraphics[width=1.3in]{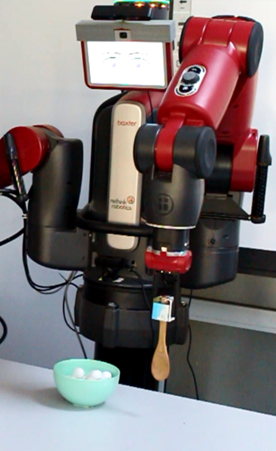}}%
\label{fig:real_environment}}
\hfil
\subfloat[]{\fbox{\includegraphics[width=1.3in]{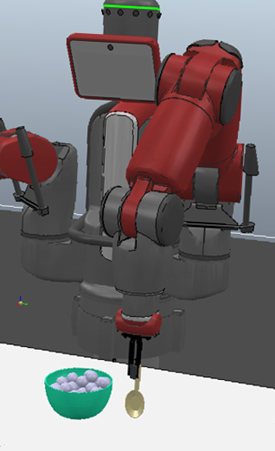}}%
\label{fig:simulated_environment}}
\caption{Experimental setups. (a) Real (b) Simulated Environment.}
\label{fig:environments}
\end{figure}

In the real world environment, a Baxter humanoid robot with a spoon attached to its gripper is situated in front of a table. A bowl ($r=8 cm$, $h=8 cm$) is placed on the table, and several white-colored, sphere-shaped particles ($r=$1$\pm$0.5 cm) are placed in the bowl. In front of the table, a Kinect One RGBD sensor is placed to track the bowl and the particles. The real environment is shown in Figure \ref{fig:real_environment}. The simulated environment is created using CoppeliaSim \cite{CoppeliaSim} and designed similar to the real environment.  40 sphere-shaped particles are placed in the bowl. The number of the particles is experimentally determined, and the best number is selected according to the competence to observe the mentioned failures. During the simulated experiments, red and blue colors are used for the particles to observe the performance of the robot on the \textit{stirring} task with the human eye. The simulated environment is shown in Figure \ref{fig:simulated_environment}.

In the simulated environment, two setups are used to obtain optimal skill policies; the fixed bowl setup and the unrestricted setup. The fixed bowl setup restricts the effect of forces from the robot to apply on the bowl, resulting bowl to keep its position and orientation. Therefore, the robot focuses on the goal without considering failures related to the bowl pose. Then, the unrestricted setup is used for learning failure prevention skills for previously unconsidered failures. Forces acting on the bowl result in failures such as sliding and overturning.

First, we present our method in the simulated environment. Details of learning the stir base skill, forming risk estimation models, and learning failure prevention skills are given in Section \ref{sec:4learning_behavior_skill}, Section \ref{sec:4risk_estimation_models}, and Section \ref{sec:4learning_prevention_skills} respectively. The skill selection among these skills and the evaluation of the proposed method that uses these skills is given in Section \ref{sec:4experimental_results}. Then, the learned skills are transferred into the real environment, and the proposed method is evaluated in the real world; whose results are presented in Section \ref{sec:real_environment_experiment}. The additional materials and videos of both simulated and real robot are available\footnote{https://air.cs.itu.edu.tr/projects/tubitak-119e436.html}.

\subsection{Learning Stir Base Skill}
\label{sec:4learning_behavior_skill}
We use the \textit{stir} skill as the base skill in our study. With the \textit{stir} base skill, the robot uses a spoon to continuously stir the particles inside a bowl without considering any potential failures. Therefore the fundamental goal of the stir base skill is to move particles as much as possible. By focusing on only one objective while omitting potential failures, we reduce the problem from multi-objective learning to single-objective learning, making the learning more straightforward.

In this work, we formulate the skill learning problem as an MDP with a tuple: $\langle S, A, T, R,\gamma \rangle$ where $s_t \in S$ is a continuous state, $a_t \in A$ is a continuous action,  $T(s_{t+1}|s_t,a_t)$ is transition probability, $R(s_t,a_t,s_{t+1})$ is the reward and $\gamma$ is the discount factor. In continuous state/action environments, Deep Deterministic Policy Gradients (DDPG)~\cite{Lillicrap2016} is one of the most prominent  approaches for skill learning. With DDPG, a skill is denoted by two policies; an actor policy $\pi$ and a critic policy $Q$. The actor policy $\pi$ maps states $s_t \in S$ to actions $a_t \in A$ in a continuous domain, and the critic policy $Q$ estimates values of state-action $s_t - a_t$ pairs. For exploration, a noise $\nu$ from an Ornstein-Uhlenbeck\cite{UOP} process is used. Acting according to $a_t$ results in a transition to state $s_{t+1}$, thus, the robot obtains a reward $r_t \in R$ depending on the task to learn. Transitions ($s_t$, $a_t$, $s_{t+1}$, $r_t$) are collected into an experience replay memory, and they are used for the optimization of the target actor and the target critic policies. Actor and critic policies are updated with target policies periodically. 

The \textit{stir} skill policy $\pi_b$ is learned as the base policy in the simulated environment. The policy has been optimized by formulating the execution as an MDP with the following state and action spaces:
\begin{equation}
\begin{split}
    S_{stir} &= \left[ x, \phi \right]~~~ \\
    A &= \left[ \Delta x  \right]~~~
\end{split}
\end{equation}
where $S_{stir}$ is composed of the position of the spoon relative to the bowl ($x$) and the phase($\phi$) of the execution, and \textit{A} is composed of the displacement of the position of the spoon ($\Delta x$). Parameters in the state and the action spaces are selected as their relevance to the problem and for simplicity, positions are represented with 2D coordinates of the plane parallel to the table. $x$ is the Cartesian positions between $\left[-\eta, \eta \right]$ where $\eta$ is the safety perimeter for the robot to operate within the allowed range. $\phi$ is the phase value representing the relative time of the execution, making the system time-variant. Using $\phi$ adds a temporal aspect to the skill, and we experimentally found it useful for periodic movements such as \textit{stir} skill. It is a value between $\left[0, \phi_{max} \right]$, and updated after each movement decision with $\phi_{step}$ using:  
\begin{equation}
    \label{eq:phi}
    \phi = (\phi + \phi_{step}) \ mod \ \phi_{max}~~~
\end{equation}
\textit{Stir} base skill is expected to reflect the essence of stirring without safety concerns, reducing the complexity of the problem. Therefore, an optimal \textit{stir} base skill policy that focuses specifically on stirring is obtained with fixed bowl setup.
Stirring can be considered as the continuous movement of the particles in the bowl hence the reward is formulated as the sum of the displacement of the particles in 50ms. The reward function is given as follows:
\begin{equation}
\label{eq:reward_behavior}
\begin{split}
    R^b(s_t,a_t) &= \sum_{k=0}^{n}  r^b(x^k_t,x^k_{t+1}) ~~~,\\
r^b(x^k_t,x^k_{t+1}) &= 
    \begin{cases}
        ||x^k_t - x^k_{t+1}||_2, &\text{if} \quad  in (x^k_{t+1}, bowl) \\
        0, &\text{else} 
    \end{cases}
\end{split}
\end{equation}
where $r^b$ is the displacement of a particle between two consecutive states 50ms apart, if the particle is in the bowl and 0 otherwise. $n$ is the number of particles in the bowl which is 40 in our setup. $R^b$ is the reward of the base skill which is the sum of individual rewards for each particle. 

Both actor and critic neural networks are designed with two linear feed-forward layers with 400 and 300 neurons respectively, and with ReLU activation layer in between. Networks are trained with Deep Deterministic Policy Gradients (DDPG) ~\cite{Lillicrap2016} for 1500 episodes with 500 steps where the batch size is 128, learning rates($\alpha_a,\alpha_c$) are 0.0001 and discount factor ($\gamma$) is 0.99. For exploration, linearly decaying epsilon is used and the noise is modelled with Ornstein-Uhlenbeck~\cite{UOP} process with parameters $\mu_\nu=0,~ \sigma_\nu=1,~ \theta_\nu=0.15$.

As the result of the training, the best policy $\pi_b$ is selected as the optimal policy for the \textit{stir} base skill\ and added to $L$.

\begin{figure}[h]
\centering
\subfloat[]{\fbox{\includegraphics[height=1.2in]{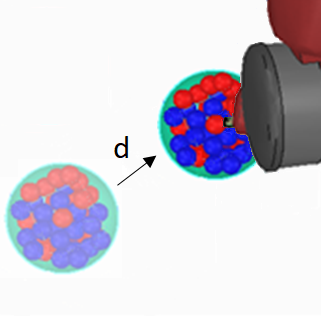}}%
\label{fig:f1}}
\hfill
\subfloat[]{\fbox{\includegraphics[height=1.2in]{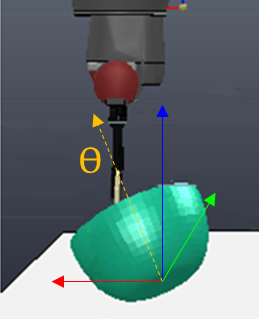}}%
\label{fig:f2}}
\hfill
\subfloat[]{\fbox{\includegraphics[height=1.2in]{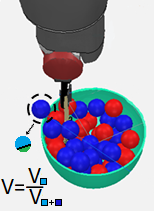}}%
\label{fig:f3}}
\caption{The observed failures during testing the learned \textit{stir} base skill. (a) Sliding the bowl (b) Overturning the bowl (c) Spilling contents from the bowl}
\label{fig:failures}
\end{figure}
\subsection{Risk Estimation Models}
\label{sec:4risk_estimation_models}
Potential failures that may occur are observed during test executions of $\pi_b$ in the simulated environment. When the \textit{stir} base skill ($\pi_b$) is tested in the fixed bowl setup, \textit{spill} failure is observed. \textit{spill} failure happens when the particles in the bowl get out of the bowl by the forces applied by the spoon. However, when the \textit{stir} base skill ($\pi_b$) is tested in the unrestricted setup, two additional failures are observed: \textit{slide} and \textit{overturn} failures. During the execution, the bowl is expected to stay close to the starting position. But forces applied on the spoon may slide the bowl away from the starting point resulting in \textit{slide} failure. Especially, the increased amount of particles results in the bowl sliding away frequently. While the bowl slides away, the friction between the bowl and the table can act as a hinge, rotating the bowl and spilling most of the particles, resulting \textit{overturn} failure. Depictions of these failures are shown in Figure \ref{fig:failures}. These failures are used as testbeds to learn failure prevention skills to augment the stir skill for safety.

For that purpose, relevant risk estimation models are designed as finite state machines(FSM) with two states, \textit{safe} and \textit{risky}. The distance between the current and the initial position of the bowl ($d$), the angle between the z-axis of the bowl and the normal vector of the table ($\theta$), and the maximum excluded volume ratio of particles ($V$) are the parameters in the design of risk estimation models for \textit{slide}, \textit{overturn}, and \textit{spill} failures, respectively. An observed parameter greater than risk activation $\kappa_a$ triggers the risk. Risk disappears when the observed parameter is reduced to risk deactivation $\kappa_d$. Designed risk estimation models for \textit{stir} base skill ($\pi_b$) and their empirically selected parameters are given in Table \ref{tbl:risk_estimation_params}.

\begin{table}[]
\centering
\caption{Risk Estimation Models}
\label{tbl:risk_estimation_params}
\begin{tabular}{|l|l|l|l|}
\hline
  & $\chi$ & $\kappa_a$ & $\kappa_d$ \\ \hline
slide & $d$ & 0.05m  & 0.02m  \\ \hline
overturn & $\theta$  & 0.3rad  & 0.1rad  \\ \hline
spill & $V$  & 0.66  & 0.33  \\ \hline
\end{tabular}
\end{table}

The risk estimation model of \textit{slide} failure uses the distance ($d$) between the initial location of the bowl ($x_0$), and the current location ($x_t$) of the bowl,  calculated as $d = \Delta(x_0, x_t)$. The risk estimation model of \textit{overturn} failure uses the rotation angle ($\theta$) between the initial pose of the bowl ($\theta^0$) and the current pose ($\theta^t$) of the bowl, calculated as $\theta = \theta^t - \theta^0$. The risk estimation model of \textit{spill} failure uses the overflown volume of particles ($Ve_n$) from the volume of the bowl ($Vb$). The ratio of the overflow of each particle ($V_n$) is calculated as the overflown volume of the particle ($Ve_n - Vb$) over the volume of the particle ($Vc_n$) as $Vn = (Ve_n - Vb)/Vc_n$. Then the overflown volume($V$) is calculated as the maximum ratio of overflow ($max(V_n)$) among the particles. The pose of the bowl and particles are directly acquired from the simulation as its ground truth value, and particle volumes are predefined.
\subsection{Learning Failure Prevention Skills}
\label{sec:4learning_prevention_skills}

Failure prevention skills are acquired by learning failure prevention policies for  corresponding risk estimation models. They are learned in the simulated environment with the same network design and optimization parameters as the learning of the base skill (Section \ref{sec:4learning_behavior_skill}).  Different from base skill learning, these skills have different state representations and reward functions. Additionally, they use initial procedures to increase the risk at the start of the episode letting the robot encounter the corresponding risk easily.

Failure prevention skills require additional parameters related to the failure they respond to on top of the base skill state representation. They use $d$, $\theta$, and $V$ parameters from the corresponding risk estimation model to learn the optimal robot movement that avoids the risk. The state of a failure prevention skill is obtained by concatenating $S_{stir}$ with $\chi$ from the corresponding risk estimation model; $d$ for \textit{slide} failure, $\theta$ for \textit{overturn} failure and $V$ for \textit{spill} failure as given in Equation \ref{eq:failure_state_reps}. The reward for each failure prevention skill is sparse and acquired with the corresponding risk estimation model as explained in Equation \ref{eq:risk_reward}. 
\begin{equation}
\begin{split}
    S_{slide} &= \left[ x_{spoon}, \phi, d \right] \\
    S_{overturn} &= \left[ x_{spoon}, \phi, \theta \right] \\
    S_{spill} &= \left[ x_{spoon}, \phi, V \right] \\
\end{split}
\label{eq:failure_state_reps}
\end{equation}
To learn a failure prevention skill effectively, risky states should be experienced during training which may not occur often enough. Initial procedures are designed to move the robot to a risky state, thus, letting the robot experience risky states. Each failure has different initial procedures. For prevention of \textit{slide}, the initial position of the bowl is sampled randomly triggering the risk $\rho^{slide}$. For prevention of \textit{overturn}, the robot moves the spoon toward a random direction until the risk $\rho^{overturn}$ is triggered. For prevention of \textit{spill}, the robot moves the spoon toward a random location in the bowl until the risk $\rho^{spill}$ is triggered. The episode starts once the corresponding risk is triggered using initial procedures, and the robot learns how to reduce the risk.

As the results of trainings, best policies of $\pi_{p_{slide}}$, $\pi_{p_{overturn}}$, $\pi_{p_{spill}}$ are selected as optimal failure prevention policies for sliding, overturning, spilling failures respectfully. Selected policies are added to the skill library forming $L_4$.

\subsection{Overall Results}
\label{sec:4experimental_results}

\begin{table*}[]
\centering
\caption{Evaluation results of augmentation of the stir skill with failure prevention skills in the simulated environment}
\label{tbl:l2_fixed_bowl_statistics}
\begin{tabular}{|lllllllll|}
\hline

\multicolumn{1}{|c|}{\multirow{2}{*}{Model}} & \multicolumn{2}{c|}{Stir Reward}                      & \multicolumn{2}{c|}{Spill}                            & \multicolumn{2}{c|}{Slide}                            & \multicolumn{2}{c|}{Overturn}   \\ \cline{2-9} 
\multicolumn{1}{|c|}{}                              & \multicolumn{1}{c|}{mean}   & \multicolumn{1}{c|}{std}   & \multicolumn{1}{c|}{mean} & \multicolumn{1}{c|}{std}  & \multicolumn{1}{l|}{mean} & \multicolumn{1}{l|}{std}  & \multicolumn{1}{l|}{mean} & std \\ 
\hline
\multicolumn{1}{|l|}{$\pi_{b}$-F}                    & \multicolumn{1}{l|}{329.00} & \multicolumn{1}{l|}{17.57} & \multicolumn{1}{l|}{\cellcolor[HTML]{e34444} 4.55} & \multicolumn{1}{l|}{1.50} & \multicolumn{1}{l|}{N/A}     & \multicolumn{1}{l|}{N/A}     & \multicolumn{1}{l|}{N/A}     & N/A    \\ \hline
\multicolumn{1}{|l|}{$\pi_{b}$-U}                     & \multicolumn{1}{l|}{251.29} & \multicolumn{1}{l|}{12.90} & \multicolumn{1}{l|}{\cellcolor[HTML]{e34444}2.2}  & \multicolumn{1}{l|}{1.32} & \multicolumn{1}{l|}{\cellcolor[HTML]{e34444}0.11} & \multicolumn{1}{l|}{0.01} & \multicolumn{1}{l|}{\cellcolor[HTML]{7ade66}0}    & 0   \\ \hline
\multicolumn{1}{|l|}{$L_4$-U}                        & \multicolumn{1}{l|}{159.64} & \multicolumn{1}{l|}{34.67} & \multicolumn{1}{l|}{\cellcolor[HTML]{7ade66}0.20} & \multicolumn{1}{l|}{0.52} & \multicolumn{1}{l|}{\cellcolor[HTML]{7ade66}0.05} & \multicolumn{1}{l|}{0.03} & \multicolumn{1}{l|}{\cellcolor[HTML]{7ade66}0}    & 0   \\ 
\hline
\multicolumn{1}{|l|}{$L_2$-F}                    & \multicolumn{1}{l|}{87.20}  & \multicolumn{1}{l|}{53.98} & \multicolumn{1}{l|}{\cellcolor[HTML]{7ade66}0.10} & \multicolumn{1}{l|}{0.30} & \multicolumn{1}{l|}{N/A}     & \multicolumn{1}{l|}{N/A}     & \multicolumn{1}{l|}{N/A}     & N/A    \\ 
\hline
\multicolumn{1}{|l|}{$\pi_{c}$-U}               & \multicolumn{1}{l|}{126.76} & \multicolumn{1}{l|}{10.70} & \multicolumn{1}{l|}{\cellcolor[HTML]{7ade66}0.15} & \multicolumn{1}{l|}{0.36} & \multicolumn{1}{l|}{\cellcolor[HTML]{7ade66}0.022} & \multicolumn{1}{l|}{0.007} & \multicolumn{1}{l|}{\cellcolor[HTML]{7ade66}0}    & 0   \\ \hline

\end{tabular}
\end{table*}
A skill library is initialized with the learned base skill $\pi_{stir}$ (Section \ref{sec:4learning_behavior_skill}). Then, learned failure prevention skills (Section \ref{sec:4learning_prevention_skills}) are added to the skill library ($L$) augmenting the base skill for safe robot manipulation. We define safe robot manipulation as skill selection in real time from the skill library consisting of a base skill to complete the task and failure prevention skills to reduce failure risks. We use a rule-based skill selection policy (Section \ref{sec:3skill_selection}). The priorities are given in Equation \ref{eq:priorities} where $I(\pi_{p_{k}})$ is the importance of the failure \textit{k} which is determined by its effect on the task. Overturning restrains completion of the task, therefore, it is the most prior failure. Spilling restrains obtaining the best outcome from a completed task, and it is set as the second most prior failure. Sliding results in soft failures such as reducing observability, and extending workspace which may lead to additional failures. In our setup, since it does not prevent the completion and the success of the task directly, it is the least prior failure.
\begin{equation}
\begin{split}
    I(\pi_{p_{overturn}}) > I(\pi_{p_{spill}}) > I(\pi_{p_{slide}}) \\
\end{split}
\label{eq:priorities}
\end{equation}
Note that, for a different setup, the importance of failures can be different from what we present. For example, if the robot stirs a pan on a stove, keeping the pan on the stove would be more important than spilling the content. 

\subsubsection{Evaluation of the Augmented Stir Skill}

The comparative results are presented in  Table \ref{tbl:l2_fixed_bowl_statistics} where all methods are tested for 20 episodes with 1000 steps. The table reports mean and standard variation of performance measures ($R,d,\theta,V$).
We first evaluate $\pi_b$ in the fixed bowl setup for benchmarking (\textit{$\pi_b$-F}). Evaluation results indicate that $\pi_b$ stirs effectively while spilling occasionally. Then, our method is evaluated in the unrestricted setup (\textit{$L_4$-U}). Comparing  \textit{$\pi_b$-F} and \textit{$L_4$-U}, we can claim that the proposed method is significant for failure prevention with a tradeoff of stir efficiency. The loss of stir efficiency is tolarable as the environment of our method is more challenging and vulnerable to failures than the former. Therefore, it uses its time effectively to prevent any of the failures and stir whenever it is safe.

For a fair comparison between our method and $\pi_b$, the latter is also tested in the unrestricted setup (\textit{$\pi_b$-U}). Comparing \textit{$\pi_b$-U} and \textit{$L_4$-U}, we see that our method is safer with a tradeoff of stir efficiency. Note that \textit{$\pi_b$-F}  does not perform better than \textit{$\pi_b$-U} for safety even though the number of spill events decreased. In the unrestricted setup, forces affecting particles get diminished since a part of the force is transferred to the bowl, causing the number of spill events to decrease. Note that no overturn event is detected in the results because overturn failure occurs when the robot interacts with the bowl which only happens with $\pi_{p_{slide}}$. While this never happens for \textit{$\pi_b$-F} and \textit{$\pi_b$-U}, \textit{$L_4$-U} successfully prevent overturn failure with $\pi_{p_{overturn}}$.
 
\subsubsection{Adaptability to Novel Failures}

To show the adaptability of our method, we show how a robot working in the fixed bowl setup adapts its library to the unrestricted setup. In the fixed bowl setup, only spill failure can be observed since the bowl is fixed, and the skill library $L_2$ is formed with $\pi_{b}$ and $\pi_{p_{spill}}$.
When the bowl orientation restrictions are removed from the environment, novel failures; \textit{sliding} and \textit{overturning} are observed, and  $\pi_{p_{slide}}$ and $\pi_{p_{overturning}}$ skills are learned to prevent them, respectively. Now the skill library is extended ($L_4$) with these skills. When we compare \textit{$L_2$-F} and \textit{$L_4$-U}, it can be seen that with our method, it is easy to adapt to new conditions by discovering novel failures, and learning corresponding failure prevention skills.

\subsubsection{Modularity vs Compound Skill}
\label{sec:4comparing_with_benchmark}

One of the main question that should be discussed is whether the modularity helps with the failure prevention problem or not. For this investigation, a compound base-failure prevention skill($\pi_{c}$) is learned that takes into account all three failures during learning to stir, and penalizes accordingly. $\pi_{c}$ is trained in the unrestricted setup with a the same training setup(\ref{sec:4learning_behavior_skill}), except the reward function. As the reward, the sum of all rewards for each objective is used as given in Equation \ref{eq:reward_bp}.
\begin{equation}
\label{eq:reward_bp}
    R^{bp} = R^b + R^{slide} + R^{overturn} + R^{spill} ~~
\end{equation}

\begin{figure}[h]
\centering
{\includegraphics[width=2.1in,height=2.0in,trim={0 0 0 1.35cm},clip]{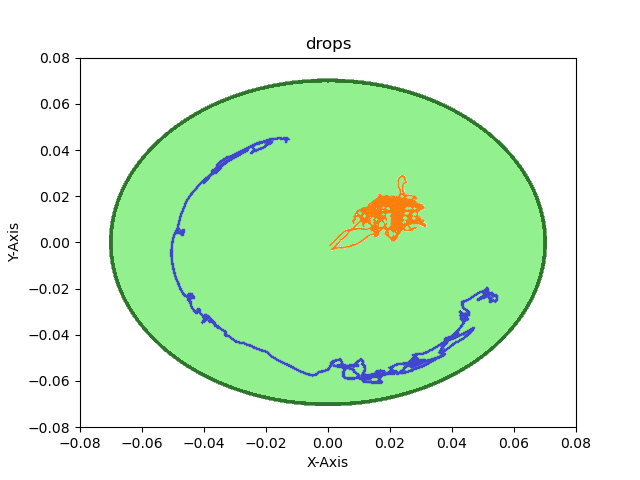}%
}
\caption{Trajectory that a particle travels using compound skill \textit{$\pi_{c}$-U}(orange) and modular method \textit{$L_{4}$-U}(blue)}
\label{fig:trajectory_tracking}
\end{figure}

Comparing \textit{$L_{4}$-U} and \textit{$\pi_{c}$-U} from Table \ref{tbl:l2_fixed_bowl_statistics},  we can deduce that our method performs slightly better for the stir efficiency and the compound skill performed slightly better for failure prevention.
However, when we compare the learned stir patterns of both methods (see a randomly selected particle's  trajectories in Figure \ref{fig:trajectory_tracking}), we see that  \textit{$\pi_{c}$-U} does not perform a circular movement, it rather moves the spoon linearly in a narrow area resulting in only slight changes in particle locations and not an effective stir. This performance degradation also supports the decrease in the observed in the average stir reward. Due to this slow pattern of movement, the probability of failures are observed are small compared to that of our method. The differences in  percentages are not significant. Our further analysis shows that modular methods reward is highly dependent on how fast the prevention policy reduces the risk.

\subsection{Transfer to the Real World}
\label{sec:real_environment_experiment}

In this work, we directly  transfer $d$ and $\theta$ parameters from simulation to the real world. However, we use domain adaptation for the rest of the parameters($V$) that can not be represented directly.

\begin{figure}[h]
\centering
\includegraphics[width=2.5in]{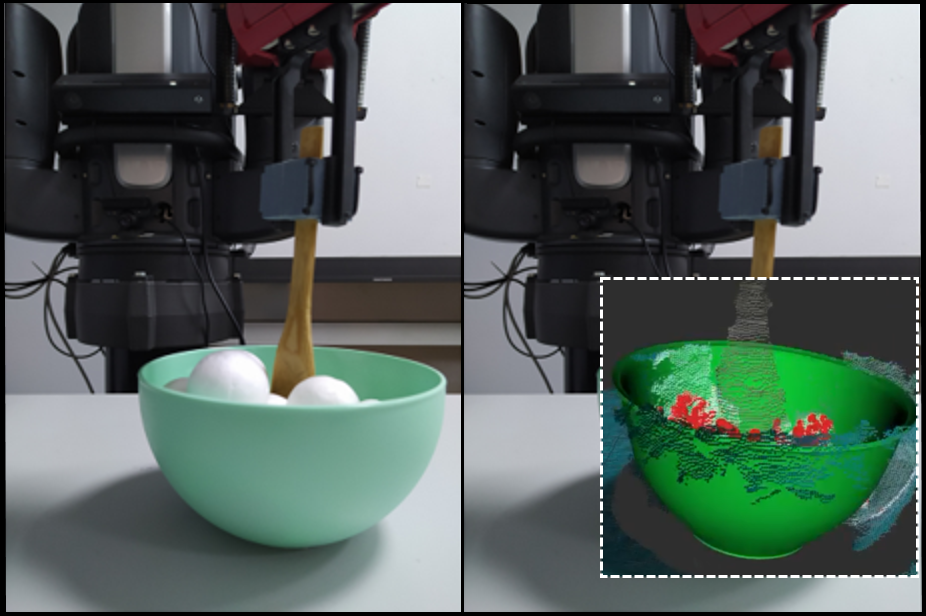}%
\caption{Visualization on the point cloud: Green is for  tracking the bowl, red is for the sampled points from the content of the bowl}
\label{fig:real_observation}
\end{figure}

For the \textit{stir} base skill, the position of the spoon $x_spoon$ which is attached to the gripper is obtained by the kinematic chain of the robot. $\phi$ is initialized as 0 and $\phi_{max}$ is set to 50. For $\pi_{p_{slide}}$, $\pi_{p_{overturn}}$, $\pi_{p_{spill}}$, the pose of the bowl is observed using a particle filter-based tracking algorithm\cite{wuthrich-iros-2013}. Then $d$ and $\theta$ are estimated from the pose of the bowl. Detecting and tracking particles in the bowl is impractical in the real world. In order to obtain $V$, we used a point cloud filtering approach using Point Cloud Library (PCL)\footnote{https://pointclouds.org}, sampling points from particles in the bowl. We estimated $max(z)$, where $z$ is the position of a sample point on the z-axis, and $max(z)$ is transferred to $V$ using Equation \ref{eq:trans_v} as the mapping function.
\begin{equation}
\begin{split}
    V \approx (max(z) - z_{bowl}) / 2r \\
\end{split}
\label{eq:trans_v}
\end{equation}
$z_{bowl}$ is the tip position of the bowl on the z-axis. $r$ is assumed to be a predefined fixed radius of a particle. The visualization of the observations from the real world is given in Figure \ref{fig:real_observation}. Action from a skill represents a desired movement vector for the gripper in the Cartesian domain. The robot is controlled using a set of position controllers in the joint state domain. We implemented a high-level controller that gets action from $L_4$ and calculates joint state goals with 100ms frequency. First,
the Cartesian goal is calculated by shifting the position of the gripper with the desired movement vector. Then, the joint state goal is calculated from the Cartesian goal using MoveIt\footnote{https://moveit.ros.org}. Then, the joint state goal is used to set desired positions of position controllers. Additionally, with 100ms frequency, $\phi$ is increased by one (Equation \ref{eq:phi}).

\begin{figure}[h]
\centering
\subfloat[]{\includegraphics[height=0.95in]{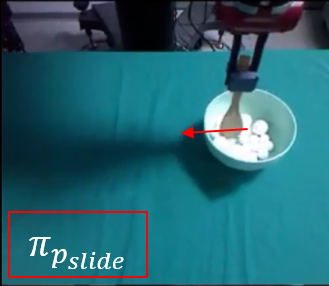}%
\label{fig:real_f1}}
\hfill
\subfloat[]{\includegraphics[height=0.95in]{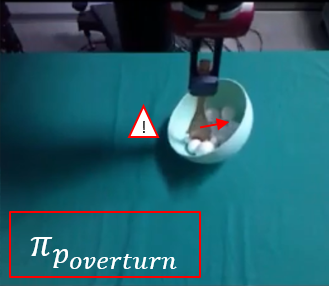}%
\label{fig:real_f2}}
\hfill
\subfloat[]{\includegraphics[height=0.95in]{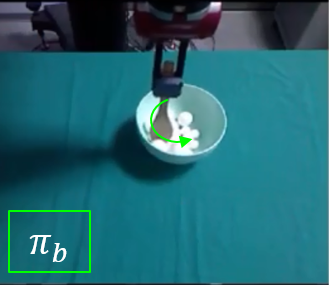}%
\label{fig:real_f3}}
\caption{An example execution trace of $L_4$ model in the real world: (a) Robot uses $\pi_{p_{slide}}$ to correct the position of the bowl (b) While executing $\pi_{p_{slide}}$, bowl starts to overturn and the robot decides executing $\pi_{p_{overturn}}$ (c) After executing failure prevention skills, the robot continues to \textit{stir} with $\pi_{b}$.}
\label{fig:example_run}
\end{figure}

By transferring $L_4$ model, we have shown our method's capability to prevent failures in the physical world. Since the exact positions of the particles in the bowl cannot be observed directly as in the simulation, the cumulative reward of the execution can not be determined in the real world. Therefore, the model's success can be stated qualitatively. Based on our observations, we can conclude that continuous \textit{stirring} with preventative skills can be successfully applied. An example execution trace is given in Figure \ref{fig:example_run}. 

We conducted additional tests to analyze the performance of $L_4$ model's preventative ability. At first, $L_4$ model is tested for each failure individually by starting the scenario in a risky condition. In these tests, the robot easily avoided failure and continued with the base skill after the risk is reduced. Then $L_4$ model is tested for 2-minute long executions. In these tests, different amounts of particles are used. When the bowl is about 70\% or less full, the robot runs as expected, prevents failures in risky states, and stirs continuously in safe states. When the bowl is more than 70\% full, the robot can not reduce the $max(z)$ easily and gets stuck in failure prevention. In some cases with high amount of particles, the actions of the robot can result in some particles popping out of the bowl. This failure is not observable by the available risk estimation models, yet may be estimated by using additional sensors such as force sensors which are not available in this work. Additionally,  $L_4$ model is tested with interference by the operator. The operator interferes the execution by changing the location of the bowl, adding additional particles, and removing particles from the bowl. The model adapts robustly to new conditions without stopping and preventing possible failures.

\section{Conclusion and Future Work}

In this work, we propose a modular method where base skills and failure prevention skills are combined. Failure prevention skills are learned for preventing potential failures and used for augmentation of base skills to make them safer by using rule-based skill selection. Evaluation results indicate that learned failure prevention skills help base skills to complete their tasks safely. The method is also extendable upon novel failures. We also transferred learned skills to be used in the real world successfully. Additional materials, and videos of simulated and real results are available online. Even though potential failures are determined using base skills, failure prevention skills are learned independently and whether they can be used for augmenting different base skills is a question that we want to address in the near future. 


\bibliographystyle{IEEEtran}
\bibliography{references.bib}
\end{document}